\documentclass[a4paper,10pt]{article}
\usepackage[utf8]{inputenc}
\usepackage[T1]{fontenc}
\usepackage{tabularx}
\usepackage{lineno}
\usepackage{graphicx}
\usepackage{times}
\usepackage{hyperref}
\usepackage{subcaption}
\usepackage{listings}
\usepackage{amsmath}
\usepackage{xcolor} 
\usepackage{hyperref}
\usepackage{float}

\usepackage[a4paper, left=1.5cm, right=1.5cm, top=2cm, bottom=2cm]{geometry}

\definecolor{codebg}{rgb}{0.95,0.95,0.95}
\definecolor{keywords}{rgb}{0.0,0.0,0.6}
\definecolor{comments}{rgb}{0.0,0.5,0.0}
\definecolor{strings}{rgb}{0.6,0.0,0.0}

\title{User-centric evaluation of explainability of AI with and for humans: a comprehensive empirical study}

\author{
    Szymon Bobek\textsuperscript{1,a} \and
    Paloma Korycińska\textsuperscript{2,b} \and
    Monika Krakowska\textsuperscript{3,b} \and
    Maciej Mozolewski\textsuperscript{4,a} \and
    Dorota Rak\textsuperscript{5,b} \and
    Magdalena Zych\textsuperscript{6,b} \and
    Magdalena Wójcik\textsuperscript{7,b} \and
    Grzegorz J. Nalepa\textsuperscript{8,a}
}

\date{}

\begin{document}
\maketitle

\begin{center}
    \textsuperscript{a}Jagiellonian Human-Centered AI Lab, Mark Kac Center for Complex Systems Research, Institute of Applied Computer Science, Jagiellonian University, Krakow, Poland \\
    \textsuperscript{b}Institute of Information Studies, Faculty of Management and Social Communication, Jagiellonian University, Krakow, Poland
\end{center}

\footnotetext[1]{Email: \texttt{szymon.bobek@uj.edu.pl}, ORCID: 0000-0002-6350-8405}
\footnotetext[2]{Email: \texttt{paloma.korycinska@uj.edu.pl}, ORCID: 0000-0002-4010-079X}
\footnotetext[3]{Email: \texttt{monika.krakowska@uj.edu.pl}, ORCID: 0000-0002-2724-9880}
\footnotetext[4]{Email: \texttt{m.mozolewski@doctoral.uj.edu.pl}, ORCID: 0000-0003-4227-3894}
\footnotetext[5]{Email: \texttt{dorota.rak@uj.edu.pl}, ORCID: 0000-0001-8113-9132}
\footnotetext[6]{Email: \texttt{magdalena.zych@uj.edu.pl}, ORCID: 0000-0001-9770-3674}
\footnotetext[7]{Email: \texttt{magda.wojcik@uj.edu.pl}, ORCID: 0000-0001-5059-858X}
\footnotetext[8]{Email: \texttt{grzegorz.j.nalepa@uj.edu.pl}, ORCID: 0000-0002-8182-4225}

\begin{abstract}
This study is located in the Human-Centered Artificial Intelligence (HCAI) and focuses on the results of a user-centered assessment of commonly used eXplainable Artificial Intelligence (XAI) algorithms, specifically investigating how humans understand and interact with the explanations provided by these algorithms. 
To achieve this, we employed a multi-disciplinary approach that included state-of-the-art research methods from social sciences to measure the comprehensibility of explanations generated by a state-of-the-art lachine learning model, specifically the Gradient Boosting Classifier (XGBClassifier).
We conducted an extensive empirical user study involving interviews with 39 participants from three different groups, each with varying expertise in data science, data visualization, and domain-specific knowledge related to the dataset used for training the machine learning model. 
Participants were asked a series of questions to assess their understanding of the model's explanations.
To ensure replicability, we built the model using a publicly available dataset from the UC Irvine Machine Learning Repository, focusing on edible and non-edible mushrooms. 
Our findings reveal limitations in existing XAI methods and confirm the need for new design principles and evaluation techniques that address the specific information needs and user perspectives of different classes of AI stakeholders.
We believe that the results of our research and the cross-disciplinary methodology we developed can be successfully adapted to various data types and user profiles, thus promoting dialogue and address opportunities in HCAI research.
To support this, we are making the data resulting from our study publicly available.
\end{abstract}

\section{Introduction}

Human-Centered Artificial Intelligence (HCAI) is a new multi- and cross-disciplinary field of AI that shifts the focus from purely engineering intelligent systems to developing and applying AI technologies aimed at broadly enhancing human capabilities. 
On the other hand these new design principles should be inline with the methods of Explainable Artificial Intelligence (XAI) thus ensuring that intelligent systems are transparent, equitable, trustworthy, meeting human needs, values, and finally, remain under human control. 
HCAI approaches aim at involving a possibly broad scope of stakeholders, including researchers, developers, policymakers, and end-users.
This assures consideration of multiple perspectives in the creation, design and implementation of AI systems, and also includes comprehensive assessment of multifaceted impacts from a user-centric viewpoint. 

As explanations are crucial for HCAI, Explainable Artificial Intelligence (XAI) has been one of the most extensively developed and researched areas of AI.
It has originally emerged after the Defense Advanced Research Projects Agency (DARPA) released a challenge~\cite{darpa}, which constituted the requirements for every automated decision-making algorithm to allow human understanding of its operations.
Shortly after the DARPA challenge, the European Union (EU) has made explainability a requirement for all AI systems through such regulations as GDPR~\cite{goodman2016regulations}, or the more recent EU AI ACT~\cite{aiact2022hacker}.
One of the major postulates of these initiatives was to "Enable human users to understand, appropriately trust, and effectively manage the emerging generation of artificially intelligent partners"~\cite{darpa}.

Therefore, one of the main \emph{challenges} in achieving adequate measure of success of explanations is to include the subjective perspective of the main beneficiary of this act, who is the explanation addressee -- the human.
Even technically correct explanation, with high values of all available quality criteria, will be not useful in case when it is not comprehended by the user or stakeholder they are designed for. 
In fact, one of the fundamental pitfalls of the current XAI techniques is that they are designed from a perspective of data scientists who are creators of the subjects they try to explain.
This creates a bias that can influence the final form and quality of explanations.

Recently, many researchers active in the area of XAI have emphasized this and other challenges in the \emph{XAI 2.0 Manifesto}~\cite{Longo_2024}.

In our work we aim at tackling the above mentioned challenges following Miller's~\cite{miller2019social} findings regarding the insights from social sciences.
It is reasonable to assert that the majority of efforts in XAI rely solely on researchers' intuition regarding what qualifies as a 'good' explanation, while in fact it may be of very low or non value to the actual addressee, who is lacking domain knowledge, or data-science background.
This has serious consequences in areas such as medicine, industry, law, and the economy, where incorrectly motivated decisions can cost human lives or large amounts of money.
In fact, the lack of comprehensibility of state-of-the-art XAI algorithms and improper adjustment of their output to specific needs of different caused many attempts to practical implementation of AI systems to fail.
The optimism about making ML/AI systems understandable for people diminished when many real-world tests and expert evaluations found that explanations often made things more confusing instead of clearer.
This has been noted by communities in almost each of the aforementioned high-risk areas~\cite{ghassemi2021xaimedpitfalls,verma2021industryxaipitfalls,roski2021xaimedfail,evans2022xaifailpathology} but finally also by the creators of XAI systems~\cite{ehsan2021xaipitfalls,molnar2022pitfalls}.

Therefore, we have been following a \emph{multidisciplinary approach} to the evaluation of XAI, combining the computer science perspective on the development of explanations for AI systems with evaluation methods from social sciences to assess how well these explanations are understood by users.

These capabilities for understanding are fundamentally important for the relevance of information and the user’s grasp in effective decision-making~\cite{xu2006relevance,lee2012gratification}.

Taking all of the above into consideration, we formulated a \emph{hypothesis} that while the existing XAI algorithms allow the addresses of the explanations to get additional knowledge about the operations of ML models, this knowledge is insufficient to get the full cognitive access to the AI model's output, and furthermore it varies across different types of addressees with different informational capabilities and domain knowledge.

To substantiate our hypothesis with robust arguments, in this paper we devised \emph {an original approach} %

for a thorough evaluation of state-of-the-art XAI algorithms, with a primary emphasis on gauging user understanding of the explanations across diverse groups. 
These groups were selected based on their proficiency in data science, data visualization, and their expert knowledge in the domain of the dataset used to train the ML model.

Our findings demonstrate that the use of methods and techniques from the social sciences, along with the inclusion of users in the evaluation of XAI, provided unique insights into the actual usefulness of these explanations for people without a technical background, which would be difficult to achieve otherwise. 
Furthermore, this study can serve as valuable input for teams focused on designing interfaces for XAI, knowledge mediation in AI systems, and human-computer interactions by providing insights into how different types of knowledge presentation and visualization are utilized by various users, depending on their background knowledge and data analysis literacy.

In this context, explanations generated by XAI algorithms are a \emph{heuristic aid}\footnote{The adjective \textit{heuristic} refers here to its basic etymology, common to all disciplines of science, i.e. $\varepsilon\upsilon\rho\acute{\iota}\sigma\kappa\omega$ [heuriskō] = to find mentally, discover, detect, comprehend, recognize, and, in the present study, designated the ability of XAI explanations to make users reconstitute the AI model operation mode.} enabling users to get a cognitive access to the content generated by AI models. 
However, the proposed explanations played substantially different roles for different user groups.
They were treated by non-specialists as a potential source of knowledge that provides sole, yet mostly insufficient access to the AI model’s output, while by experts as an opportunity to compare their expertise with AI, which ended with critical assessment of the compliance of AI models with real life scenarios.
The comments of non-expert participants gave us an insight into the plethora of practical tips concerning further improvements of the usability of AI explanation methods, while the comments of experts about the quality of the dataset itself. 

The remaining structure of the paper follows the guidelines of the journal, including the presentation of specific research results, their in depth discussion, and finally methods used to obtain them.

\section{Results}
The study focused on evaluating comprehensibility of explanations generated by XAI algorithms for a ML model for mushroom classification to distinguish edible and poisonous ones.
The selection of the dataset followed \emph{two requirements} we formulated: 1) an access to experts in the dataset domain, 2) research reproducibility.
To this goal we decided to use a dataset originating from the UC Irvine Machine Learning Repository\footnote{\url{https://archive.ics.uci.edu/dataset/848/secondary+mushroom+dataset}}.
The ability to differentiate between edible and poisonous mushrooms which are traditionally and culturally harvested in large quantities for food and commercial purposes in Poland, was deliberately selected as the domain. 
Not only this is a well understood domain by a larger audience, but also, using our research and community channels we had access to high quality experts in mycology.

Using the selected dataset we trained a Gradient Boosting Classifier (XGBoost) \emph{model} achieving an accuracy rate of 99.97\
For this model we generated \emph{explanations} with some most commonly used XAI algorithms and shown to the participants of the study.
To cover the largest possible spectrum of the XAI methods and not overwhelm participants with their variety, we selected following representative types of explanations:
1) statistical descriptions and visualization of data, such as table presenting descriptive statistics (mean, median, standard deviation, quartiles), information on missing data for each variable and plots illustrating frequencies of various features
2) feature importance attribution-based explanations such as SHAP and LIME~\cite{SHAPlun,lime},
3) rule-based explanations such as Anchor~\cite{anchors}, and finally
4) counterfactual explanations such as DICE~\cite{dice2020}.
Additionally both feature importance attribution explanations as well as rule-based explanations were provided in a form of local and global explanations~\cite{FromLocalExplanationsToGlobal}.
In a selection of methods we followed loosely the work of Baniecki et.al~\cite{baniecki2023grammar} which demonstrated the usefulness of sequential analysis of a model as a combination of multiple complementary XAI mechanisms.
We presented the aforementioned explanations to the participants in the form of a \emph{set of slides}. 
We encourage readers to review the supplementary materials accompanying this paper, which include the aforementioned slides, to facilitate understanding of the subsequent analysis.

The experimental group for the study was formed of \emph{different types} of participants. 
From the \emph{perspective of expertise} in the domain of macrofungi, we could distinguish two groups: 1) mycologists and mycophiles (called hereafter “domain experts”, or \emph{DE}), and 2) non-specialists in mycology recruited from students from Jagiellonian University.
Both groups exhibited various levels of literacy in data analysis.
In the non-specialists group we identified two distinctive partitions of students, those with high level of familiarity with data analysis and visualization (\emph{IT}), and those with a lower one, originating from the  social sciences and humanities (\emph{SSH}).

\begin{figure}[ht]
\centering
\includegraphics[width=1\textwidth]{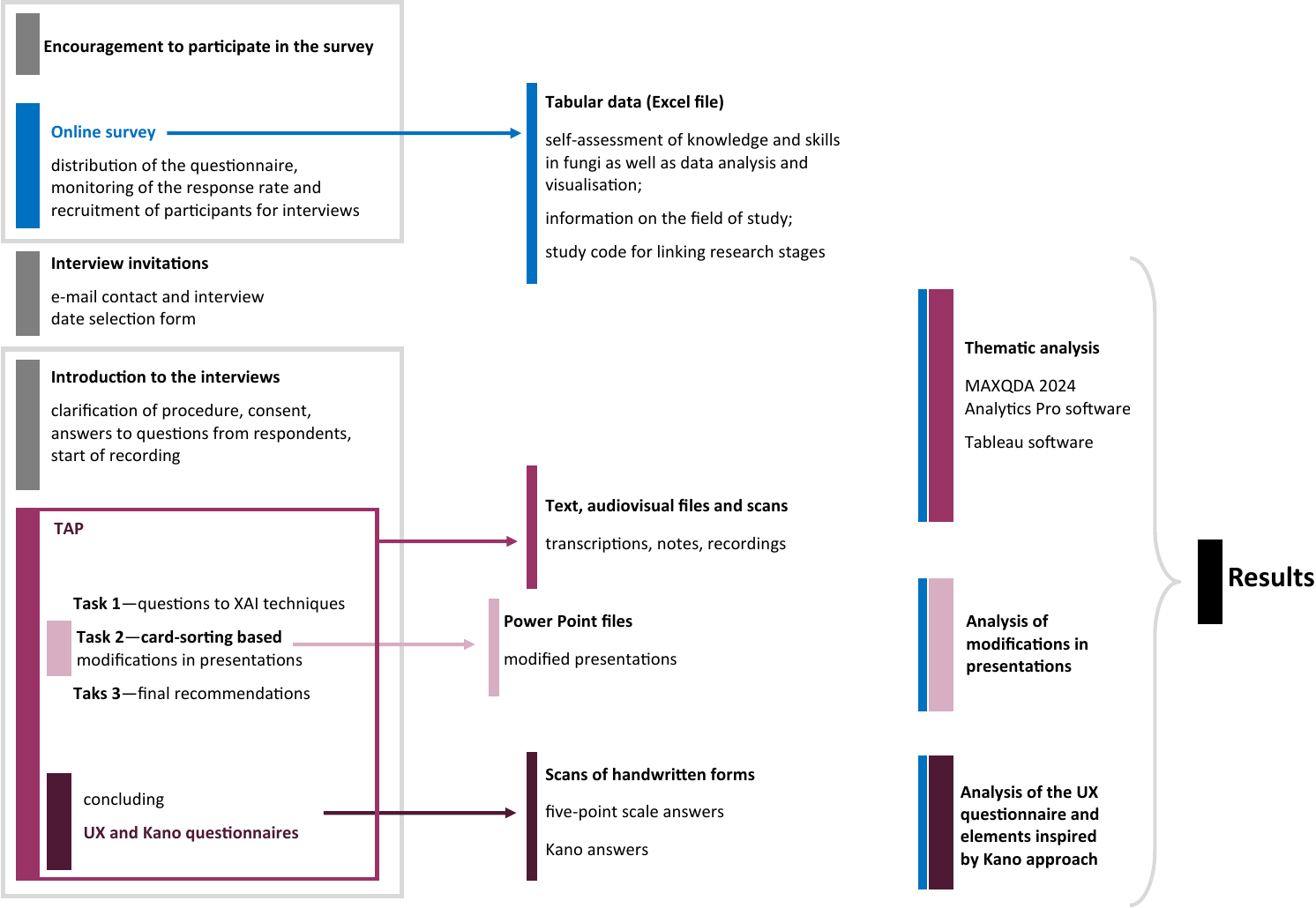}
\caption{Detailed description of the user study}
\label{fig:user-studies-details}
\end{figure}

The first part of the study with students was conducted in the Faculty of Management and Social Communication building at Jagiellonian University in Krakow, between 19 December 2023 and 25 February 2024, with data collected through an online survey (the last survey was completed on 23.02.2024) and interviews conducted using the \emph{Think-Aloud Protocol (TAP)}~\cite{nielsentap}.
The second part involving TAP interviews with mycologist and mycophiles (DE), also combined with an online survey, took the form of online video calls via MS Teams, which was convenient for respondents many of whom reside remotely from Krakow, and of spanned the period between 30 January 2023 and 4 March 2024.
The online questionnaire, TAP scenario and research tool employed in both parts of the study were identical.
Details of the user study are shown in Fig.~\ref{fig:user-studies-details} and more in-depth description of the methodological toolkit used to conduct and analyse the results is presented in Sect.~\ref{sec:methodology}.

\subsection{Thematic analysis with think-aloud protocol}

The results of thematic analysis should be interpreted in the context of a double metacognitive filter, as is standard practice in qualitative research~\cite{puryear2016inside,josephsen2017qualitative,rojas2020qualitative}. 
Furthermore, the intensity of this relevance can be indicated by aggregating and comparing the co-occurrence of codes in segments (units of thematic analysis) within individual observations (within the analysis of the empirical material produced during the study of a given user).
The first metacognitive filter, which was included in the study, pertains to the verbalisation of thoughts constructed by the user.
In contrast, the second metacognitive filter, which was also included in the study, refers to the coding of empirical material by the researcher.
In the context of the application of this dual metacognitive filter, the occurrence and co-occurrence of codes for a given user indicates the meaning of the codes in question and the links between them for the user.
Here, meaning is understood as the user's preoccupation with a given topic, object or problem, or task.

Based of this,
between January 17, 2024, and March 4, 2024, a total of 66 items were collected for coding, including respectively: (a) 13 transcripts of research video sessions with domain experts, (b) 27 transcripts of study conducted with students and 26 notes prepared by the think-aloud protocol facilitators after the study's conclusion.
For a detailed list of codes used in the coding of collected items, please refer to the supplementary materials.

In the IT group, a subgroup of students, there were more frequent codes related to the following criteria: appropriateness, representativeness, comprehensibility, but also unreadability. 
Unsurprisingly, there were more references to knowledge and experience in visualisations and analysis but also in analysing human behaviour, which is interesting especially as the second group included, among others, students from psychology. 
Perhaps the frequent occurrence of this code indicates IT students' interest in user experience issues, but this claim would need to be verified in the form of a separate study. Following this pattern, codes relating to barriers, difficulties, discrepancies, as well as issues relating to expectations, trust are also more frequent among IT students. 
Affective elements were also noted - expected interest on the one hand, but also more frequent overwhelming and confusion on the other. 
The IT student group also proved to be richer sources of codes concerning associations and personification.

In the SSH group codes relating to aesthetics, including colour scheme, were more frequent. This group also provided more recommendations and comments on the addition of supplementary materials, reference to graphics, feature significance and the usefulness of coordinate axes. Codes on the following evaluation criteria were more frequent: credibility, efficiency, importance, but also uselessness. In addition, more references to knowledge and experience in macrofungi were identified in this group, as well as the following affects: uncertainty, tiredness, reluctance and nervousness.

Regarding the DE group, in order to keep this recapitulation concise and synthetic, we only recorded the findings that show a clear contrast with both groups of students. When comparing the occurrence rate per observation calculated for respective codes and summed up, 
what strikes in the first place is the presence of significant discrepancies noted for several rubrics. With 100\
The dominant sentiment, which was also much more frequent than among students, was overwhelming (100\

Another noteworthy difference between students (SSH and IT) and experts (DE) is revealed in the prevalence of mentions framing with the code “Method of analysis” (100\
This uniform attitude was observed consistently across all experts, with no deviations noted. The high ratio of the "Methods of analysis" code correlates with equally high scores of the subcodes assigned to this category, which were created for experts only. Thus, during TAP interviews, in response to the main question: “What does a given visualization tell you about how to distinguish edible mushrooms from inedible or poisonous ones?”, all experts concurred that the toxicity of a fungus is contingent upon its species. This is reflected in the maximal hit rate observed for the "taxa" subcode (100\

A singular behavior, exclusively observed in experts and indicative of their cognitive perspective, which diverges from that of students, is "omission." This term was employed to thematize sequences of experts’ storytelling wherein they entirely abstained from interpreting the content of the slides, substituting it with explanations about the biology, and systematics of macrofungi. Two reasons for the occurrence of such hiatuses were identified during the analysis of the material. (1) The expert is unable to comprehend the visualization, and in a coping attempt to compensate for a perceived deficiency in his/her own skills, instead of responding to the researcher's queries regarding XAI charts, she/he displays her/his mycological knowledge, which is not pertinent to the subject matter. (2) The expert declines to interpret the XAI content and respond to the researcher's inquiries, citing her/his own mycological expertise to argue that the data presented in the charts are of no value in identifying the species and, consequently, in determining the toxicity of the fungus. Furthermore, the experts' interpretations included numerous, albeit succinct assumptions about (1) how the AI model did assign fruiting bodies to species and (2) how the model could do it better if the entry dataset set had been more adequately supplemented. These threads were coded as "AI hypotheses" and occurred in 76,90\

With regard to other coding themes are concerned, the experts scored higher than students in recommendations, expectations, several evaluation criteria (importance, efficiency, representativeness) and layout. The notable discrepancy in expectations (92,30\
 
Among students, references to knowledge and experience in macrofungi were made most frequently in reference to LIME, waterfall chart and counterfactual analysis. 
These were most often accompanied by sentiments such as surprise and uncertainty. 
In the setting of experience and knowledge in macrofungi, there were also frequent references to trust in AI and the appropriateness criterion. 
However, the codes mentioned were also linked to associations regarding the image of the toadstool and difficulties experienced when analysing XAI techniques. 
Knowledge and experience in visualisations and analysis were less strongly associated with the other codes from the code book. 
Among the stronger associations were only references to the comprehensibility criterion and box plots.

In the group of experts, the table of co-occurrence of codes pertaining to knowledge with other codes illustrates, once again, the respondents' predominant metacognitive perspective, with XAI explanations being interpreted primarily through the lens of mycological knowledge. 
Indeed, this variable overlapped extensively with all the examined XAI techniques and other elements of the research tool, with the highest number of cross-references observed for waterfall and bee swarm visualisations, followed closely by anchor, histograms, distribution of features, text slides, counterfactual analysis, LIME, box plots and descriptive statistics. 
The analysis also demonstrated a significant interconnection between KNF and the majority of methods of analysis, with the following order of priority (with only minor differences): taxa, identification practice, AI hypotheses, omission, reflection time, difficulties and thoroughness. 
With regard to the evaluation criteria, there was considerable interrelation between KEF and comprehensibility, importance, unreadability, appropriateness, precision, efficiency and representativeness. 
Moreover, it was also intricately liked with trust and, in the realm of affects, it cooccurred the most frequently with uncertainty and confusion. 
A notable connection was observed between knowledge and experience in macrofungi and expectations regarding the enrichment and profiling of the future dataset on which the ML model would be trained, as well as with barriers (including concerns) and positive reception. 
In the rubric of data features, mycological knowledge intersected with almost equal frequency with lack of significant features, significance of features, understanding of annotations, understanding of data gaps, lack of appropriate data and legend. 
With respect to knowledge and experience in analysing human behaviour, the same high frequency of overlapping was evidenced for: (1) anchor, LIME, SHAP-related techniques (waterfall, bee swarm), text slides, box plots, histograms, (2) identification practice and taxa (overarching code “Methods of analysis”), (3) uncertainty (“Affects”), (4) comprehensibility, unreadability, importance, (5) lack of significant features, significance of features (“Data features”). 
It is noteworthy that when offering commentary on human bahaviour, the experts referred exclusively to various conducts adopted by three discerned groups of users: mycologists, mycophiles and amateur mushroom collectors. 

Trust was most strongly associated with the criteria credibility, appropriateness, precision, importance, efficacy, comprehensibility, accuracy, as well as the two knowledge and experience types (in macrofungi and in analysing human behaviour) and the waterfall chart for SHAP and LIME techniques. 
However, in the setting of the trust code, the association with the image of a toadstool also stands out, as does in case of the code knowledge and experience in macrofungi. 
In the group comprising experts, the theme of trust coincided the most frequently with knowledge and experience in macrofungi. 
In regard to other codes, elevated rates of coocurrences were for found for: (1) waterfall chart, bee swarm plot, text slides, anchor, LIME, histograms, distribution of features, (2) expectations concerning data on biotope and macroscopic features of macrofungi, (3) two methods of analysis, i.e. taxa and identification practice, (4) appropriateness, comprehensibility, importance, unreadability, (5) lack of significant features, significance of features, (6) barriers, and (7) uncertainty. 
Please see the supplementary materials for a comprehensive list of codes.

\subsection{Reception of the presentation of XAI algorithms}

\begin{figure}[H]
\centering
\includegraphics[angle=90,height=0.95\textheight]{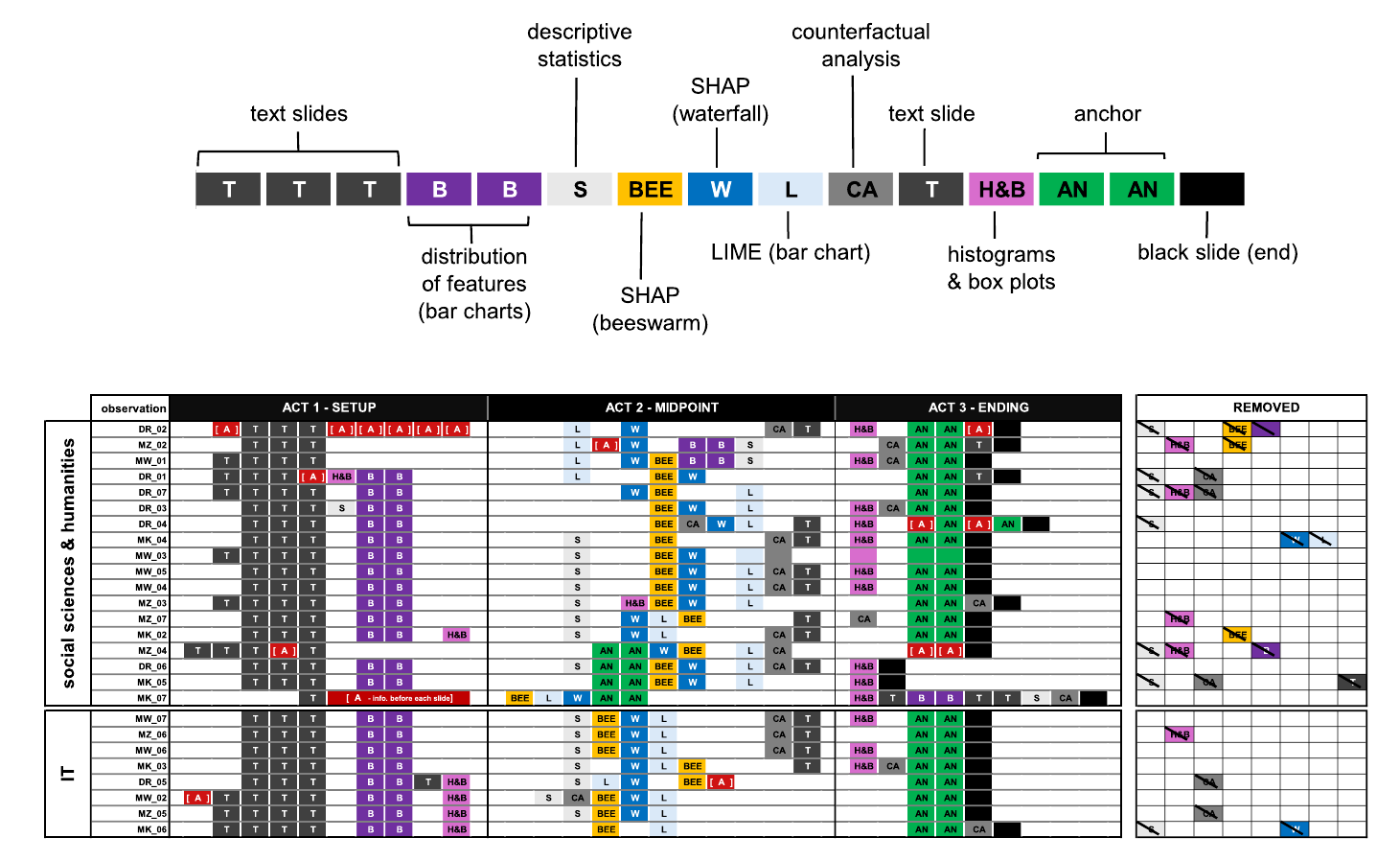}
\caption{Analysis of modifications to the order of explanations and the level of details. The original sequence of slides is presented above the table with modifications suggested by participants.}
\label{fig:order-analysis}
\end{figure}

The findings discussed in this section derive from a comprehensive synthesis of the thematic analysis, the review of modified presentations, and the Kano and UX questionnaires~\cite{Mikuli2011ACR,Laugwitz2008ConstructionAE}.
Based on the results of the thematic analysis, the majority of recommendations for the preparation of supplementary materials were related to box plots and histograms, in the context of understanding of annotations, legends and feature significance, among others.
Conversely, requests for changes of visualisation scheme were mainly related to descriptive statistics and bee swarm plot.
The study revealed students’ preferences regarding the construction of visual narratives and improving the reception of presentations by the audience.
Fig.~\ref{fig:order-analysis} presents a summary of all of the modifications proposed by the participants. 
Participants were asked to indicate which slides were not helpful and should be removed from the presentation by manually moving them beyond the last slide, which was a blank black screen, which is represented as the \textit{black slide (end)} in the figure. 

The presentation arrangements are divided into three parts: Act 1 (\emph{Setup}) is an introduction to the topic, solution or information product. 
Act 2 (\emph{Midpoint}) is the middle section where specific issues are presented or developed. 
Act 3 (\emph{Ending}) is the conclusion. 

The initial section of the presentation, designated as \emph{Setup}, should include a number of elements such as textual slides, such as the preferred title slide, a description of mushroom features, and explanatory slides about mushroom features. 

The elements placed in the subsequent order (\emph{Midpoint}) were more varied and included different sequences of the following slides: descriptive statistics, LIME, waterfall chart, bee swarm plot, counterfactual analysis and textual slides with additional explanations (e.g. explanation of LIME or how each of the XAI techniques works – e.g. observation MK\_07). 
For the four users from the social sciences and humanities group, it was deemed necessary to relocate Anchors from the end of the presentation (as it was in the base presentation) to the middle part of the presentation. 
The rationale provided by the users for this action was a highly subjective evaluation of the readability and usefulness of the anchors in terms of how the AI classifies objects (mushrooms) into classes (edible, inedible) and the desire to present an audience-friendly solution as early as possible in the presentation.
In general anchors were viewed positively in terms of readability and usefulness in both student groups surveyed. 
This is evidenced by the fact that none of the respondents moved anchors behind the black slide. 

The majority of respondents chose to leave the anchors at the end of the presentation (Act 3), as an effective summary of the whole and the solution to the problem presented in the beginning of the presentation (Act 1). 

The conclusion of the presentation (\emph{Ending}) comprised the aforementioned anchors, accompanied by histograms and box plots, as well as supplementary explanations (e.g. the functioning of the anchor and its implications) and summaries, and on occasion, counterfactual analysis. 
In the case of one observation (MK\_07), the order of the slides differed significantly from that of the other observations. 
The majority of the text slides were placed at the end of the presentation, along with the distribution of features in the form of bar charts. 
In addition to the aforementioned techniques, the following were employed: histograms and box plots, descriptive statistics, and counterfactual analysis. 
Furthermore, anchors and bee swarm (Act 2) were placed at an early stage, while the beginning consisted of text and recommendations to explain how each technique works before presenting it. 

The analysis of the visualisation feedback reveals a nuanced perspective on the strengths and weaknesses of the presentation as perceived by the students. 
One notable positive suggestion pertained to the inclusion of explanatory elements within the presentation. 
Students emphasised the importance of providing clear explanations of how each chart functions, as well as incorporating titles for individual charts and descriptive statistics. 
These additions would serve to contextualise the data presented, making it more accessible and understandable to the audience. 
Furthermore, the students emphasised the importance of enhancing readability by including legends for charts and providing detailed explanations of the data they represent. 
This focus on providing comprehensive context and guidance demonstrates a commitment to facilitating a deeper understanding of the information presented. 
Another significant aspect of the feedback relates to the presentation's visual aesthetics. 
Students identified various opportunities for improvement, including adjustments to colour contrast, text alignment, and the highlighting of key information. 

The most crucial aspect in this context was categorising visualisations according to the type and subjectively assessed level of difficulty in interpretation. 
When comparing the results of analysis among IT and SSH students, it is notable that some important differences in the modifications suggested by these two groups are particularly evident.
From a qualitative perspective, the modifications suggested by IT students were not as significant in comparison with SSH students. 
Consequently, it was possible to observe greater similarity of their sequences compared to the base presentation. 
The analysis of the feedback on the visualisations reveals a multifaceted set of recommendations aimed at optimising the clarity, relevance, and visual appeal of the presentation. 
By incorporating the suggestions of IT and SSH students, visualisations can serve an informative function, particularly in the context of an audience that lacks experience in mushroom picking or professional knowledge about them. 

In addition to the conducted analysis of presentations modified by students, the UX questionnaire analysis helped to better understand the overall user experience connected with the reception of XAI. 
For all groups included in the study, the explanations users saw during research were perceived as rather interesting and rather aesthetic. Differences began to become apparent at the stage of assessing difficulty, readability and comprehensibility. 
Students without technical background considered the explanations they saw to be rather legible, however also rather difficult to interpret and rather incomprehensible. 
Students with technical background perceived explanations they saw as rather legible and rather understandable but at the same time moderately or rather difficult to interpret which means that they were able, at least partially, to understand explanations they saw. However, it was a moderately or rather difficult process for them.
Domain experts considered the explanations they saw to be rather legible, rather understandable and moderately difficult to interpret.
The analysis of responses inspired by the Kano approach showed, through the concept of comparison, the subjective degree of difficulty of the AI visualizations perceived by users.
For the majority of respondents in all groups, the AI explanations were considered the most difficult types of data visualization they have ever seen.
The respondents easily mentioned the easier types of data presentation than those generated by AI, e.g. visualizations shown in the media, on the Internet, infographics or visualizations in textbooks, however, they were rarely able to point out more difficult ones.
The few examples indicated as more difficult to interpret that AI explanations they saw during research included visualizations of medical procedures and specialized or scientific visualizations in the field of natural and technical sciences (physics, mathematics, logic, advanced statistics, biology).

\section{Discussion}
Referring to the research hypothesis we formulated at the beginning, we can observe that
explanations generated by XAI algorithms being treated as a heuristic aid helped users to get cognitive access to the content generated by the AI model. 

However, the evidence we have collected via our multidisciplinary and mixed method research revealed that for students who declare little knowledge of macrofungi (low self-assessed level of macrofungi literacy), generated explanations proved to provide sole, yet insufficient access to AI model’s output. 
On the contrary, for mycologists and mycophiles, explanations refer directly to the primary dataset and are used as a handy tool for assessing its quality comprehended as the adequacy of data with botanic reality of macrofungi.
Moreover, in the group, there were students who express a high level of trust towards AI, even declaring abandonment of their own proclaimed extensive knowledge of mushrooms in favour of AI, or are uncertain about the outcomes and effectiveness of AI. Within this group, there were students who express a high level of confidence in the accuracy of the dataset.
Some participants in this group express a priori belief that it was compiled with consultation from competent experts in the field.
There were even those who assumed that the data were artificially generated by ChatGPT and thus may not necessarily be true. This statement should be tempered by possible cognitive and hierarchical biases (the study was notoriously conducted by teachers on students). 
Mycologist and mycophiles manifest a purely pragmatic and utilitarian approach to the provided explanation as well as to the AI model itself and, consequently, they are, in the first place, prone to assess the dataset quality, and point out its shortcomings.  

Deeper analysis of the results revealed that for students, the explanations generated by XAI algoritms were insufficient, because strictly limited to superficial visual level of explanations.
The presented infographics did not allow them to reach beyond to the primary dataset and seek coherence between displayed data and botanic reality.
With this caveat in mind, the research with illiterate students delivered nevertheless a plethora of practical tips for further improving visual and textual XAI explanations.
These students’ suggestions, both explicit and implicit, concerned indeed every single aspect of visual aesthetics and textual content of XAI explanations as well as of the research tool in the form of slideshow presentation used to guide the TAP procedure and dataset quality.
Despite the lack of mushroom expertise, some students, although few in number, made an effort to evaluate the credibility of the dataset on which the ML model was trained, as well as the safety of using the clues embedded in the explanations for real macrofungi recognition. 
This cognitive attitude should be retained as an indicator of high intellectual maturity and high information literacy level. 
Interestingly enough, the overall level of skills in reading infographics had no significant incidence on the results of interpreting XAI explanations,  
Insofar as, for students (with rare exceptions), explanations mainly refer to themselves, we may postulate that, for macrofungi illiterate recipients, infographics generated by the given AI model are confined to an autotelic function.

For the group of domain experts participants, explanation and its informative potential are secondary and subaltern to the completeness and botanic relevance of the dataset on which the AI model has operated.
When they detect a discrepancy between the content of an infographic and their mycological know-how, they simply abandon any further evaluation of the infographic itself, 
For mycologists and mycophiles, explanations have a proper mediation function, in the sense that they are perceived as a gateway giving insight to the dataset and AI model’s operation rules; the autotelic aspect of explanation is absent. 

\begin{figure}[ht]
\centering
\includegraphics[width=.5\textwidth]{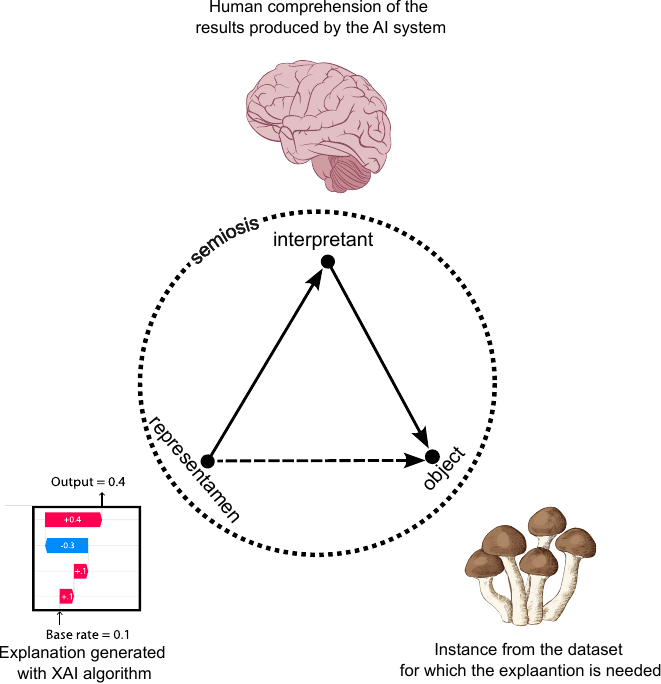}
\caption{Peirce’s semiotic triangle and its XAI interpretation with \textit{object} related to instance that is being explained (mushroom in our case), representamen related to the explanation itself (e.g. SHAP values, LIME, etc.)  and interpretant representing the comprehension of the sign (i.e the results produced by the AI system) by the user. }
\label{fig:semiotic-triangle}
\end{figure}

We approached the analysis of comprehensibility of XAI output by adapting the semiotic triangle that originally was proposed as a model that explains how linguistic symbols relate to the objects they represent. This account, originally conceived by C. S. Peirce, provides a valuable framework for understanding signs and their meanings through the triadic relationship of the following components: symbols (representamens), objects and interpretants as depicted in Fig.~\ref{fig:semiotic-triangle}.
In this work, we adapt this approach to XAI area to investigate the interplay between model outputs, input data, and human understanding~\cite{atkin2023pierce}.

In the context of explanations provided by XAI methods, the symbol represents the model output or the prediction made by an AI system. It could be a classification label, a regression value, or any other form of output generated by the model. The symbol (\emph{representamen}) is the observable result produced by the AI system, which may be correct or incorrect. The object from the perspective of XAI corresponds to the real-world phenomenon or the data instance that the AI system is supposed to predict. It could be an image, a text document, a medical diagnosis, or any other input data. The object is what the AI system aims to represent or interpret through its predictions. The interpretant refers to the meaning or understanding that arises from the interaction between the symbol and the object. In XAI, this involves human interpretation, i.e. how users (such as data scientists, domain experts, or end-users) understand and make sense of the model’s output and explanation methods, i.e. techniques that generate explanations to bridge the gap between the sign (model output) and the object (input data). 

This fundamental framework was used by us to analyze and conclude results of our study. With regard to the XAI techniques under examination, we define comprehensibility as the property that enables users to complete a full semiotic circle, moving back and forth between the following three stages: (1) perceiving and deciphering a symbol (explicit XAI content), (2) identifying the chain of objects underlying this symbol (ML algorithm operation mode <= input data <= specimens of macrofungi), and (3) embracing and integrating both of the later into a mental representation Accordingly, comprehensibility is impaired when at least one of these steps is not completed by the user.    

Following the  basic concept of Peirce’s semiotic triangle introduced above,
we can conclude that the semiotic triangle is incomplete for all of the participants groups. 
Mycologists and mycophiles bypassed the interpretant and, by a semiotic shortcut, used the representamen to reach directly to the object/referent, i.e. real macrofungi taxonomy.
They skip the interpretant and, subsequently, do not even try to reach comprehension of how the AI model engendered the displayed results. 

Student, regardless of the visual reading skills, were stuck in the relation between representamen and interpretant, failing to obtain both full understanding of the AI model operating mode and the congruency between the dataset and biological evidence on macrofungi edibility/inedibility or toxicity.
They attempt to decipher the representamen which, when disconnected from the object, fails to generate meaning. There is a blind loop between representamen and interpretant, which refer to and mirror only each other, ignoring the object. Autotelic aspect resides precisely in omitting the object, as if representamens existed solely for their own purpose.

The outcome of our research reveals how XAI explanations are perceived differently by various stakeholder groups,  highlights the need to design explanations that accommodate diverse audiences, fostering inclusivity in AI development in line with the HCAI paradigm. 
Our findings lead us to conclude that developing XAI systems aimed at automating decisions while also enhancing and supporting human abilities is feasible only if we carefully consider the needs and comprehension capabilities of all stakeholders. 
Achieving this requires the adoption of innovative human-based evaluation strategies that will lay a strong foundation for assessing XAI methods through empirical research and help identify effective approaches for improving and measuring understandability.
Research in this area should serve as a catalyst for advancing human-centered XAI, which was one of the primary aims of this study. 
We are confident that the rich empirical material collected through interviews and rigorously analyzed throughout our research will make a significant contribution to this goal.

\section{Methods}
\label{sec:methodology}

\subsection{Dataset, AI model and XAI algorithms}

The dataset we used comprises data on 61,069 specimens from 173 mushroom species, categorized as \textit{edible} or \textit{inedible/poisonous}. 
Specimens with unknown edibility were classified as inedible/poisonous. 
The dataset exclusively contains cap-and-stalk mushrooms with gill hymenophores. 
It includes both real observations and hypothetical data, the latter being artificially generated from a smaller set of real-world mushroom observations. 

Regarding class distribution, 33,888 instances are inedible or poisonous, and 27,181 are edible, making the dataset fairly balanced with approximately 55.49\
This balance is essential for accurate model training and performance evaluation, avoiding biases towards the majority class. 

The 
ML model which was later elucidated to the study participants is the Gradient Boosting Classifier (XGBClassifier). 
To handle categorical variables, a one-hot encoder was used, and missing data were addressed through imputation. 
Additionally, numerical variables were scaled to ensure uniformity in data handling. 
This preprocessing involved imputing missing values in numeric features with the median and scaling these features, while categorical variables were imputed with a placeholder value '\_NA\_'. 
This resulted in the model operating on 82 features derived from the original 20 input variables. 
Renowned for its efficiency in classification tasks, the model exhibited an accuracy rate of 99.97\
Study participants were informed about the model's limitations, emphasizing that while the model is data-driven and highly accurate, it does not encompass all possible factors affecting mushroom edibility and thus should be used as a supportive tool rather than a sole determinant in the identification process. 
Additionally, the standard practice of data partitioning was applied, segregating the dataset into a training set for developing the model and a test set for its validation. 

The explanations were generated with known public Python implementations of the methods selected for the study.
In particular this included LIME\footnote{\url{https://pypi.org/project/lime/}}, SHAP\footnote{\url{https://pypi.org/project/shap/}}, Anchor\footnote{\url{https://pypi.org/project/anchor-exp/}}, DICE\footnote{\url{https://pypi.org/project/dice-ml/}}. 
Statistical analysis were generated with the scikit-learn and matplotlib libraries\footnote{\url{https://pypi.org/project/scikit-learn/}, \url{https://pypi.org/project/matplotlib/}}.

\subsection{Participants}
The participants selected for the interviews were recruited based on the initial survey.
The survey comprised 18 questions, 12 of which pertained directly to self-assessing knowledge and skills in mushrooming and data visualisation, as well as acquiring information about their origin and experience. The survey included a question about the field of study to ensure appropriate representation of students pursuing humanities, social sciences, and computer science. Five questions in the questionnaire directly related to giving consent for data use and participation in further research, as well as confirming familiarity with the GDPR clause. The questionnaire aimed to gather information and assess the knowledge, qualifications, experience, and competences of the participants. It consisted of six open-ended questions and six closed questions. The latter included an 'other' option, allowing participants to provide their own answer. 
In total 143 respondents completed the questionnaire, including 79 students who provided consent for further participation in the study. 
Three respondents did not receive invitations 
as they did not meet the qualification criteria due to being from different academic disciplines, possessing varying levels of education, etc.

The experimental group was formed of different types of participants. From the perspective of expertise in the domain of macrofungi, we could distinguish two groups: 1) mycologists and mycophiles (called hereafter “domain experts”) and 2) non-specialists in mycology. Both groups exhibited various levels of literacy in data analysis. 

The group of domain experts included: a couple of entrepreneurs owing a laboratory-based medicinal and exotic macrofungi mycelium production plant (offering mother cultures on Petri dish, mother spawn on grain or sawdust, plug spawns and growkits), one academic scholar in chemistry, specialized in mycology, one doctoral student in mycology, two professional certified phytologists-mycologists involved in natural sciences education and macrofungi knowledge vulgarisation, four amateur mycologists, among whom one doctoral student in medical science, co-administering or moderating two of the most influential and well-reputed social media groups devoted to macrofungi identification, two amateur mycophiles holding, each of them, a popular fungal thematic channel on general-public webstreaming platforms, ans one amateur mycologist collaborating with academic mycologists in identifying and describing new fungal taxa (preparation of dried specimens for scholars). The majority of respondents are also certified macrofungi identifiers accredited by competent Polish Regional Office of Sanitary and Epidemiological Vigilance (SANEPID). This group brings together representatives of various professions with unequal levels of infographic reading competences.    

The group of non-specialists in mycology included students from Jagiellonian University representing various fields of science, social sciences, humanities, and computer science, with varying levels of information literacy skills, including different abilities in reading infographics. Students participating in the study also reported various levels of knowledge about macrofungi and experiences in their identification and assessment of edibility, ranging from decidedly low, low, moderate, to high. Despite declarations of very high competencies in fungal knowledge, as well as communicated extensive experience in identifying edible, inedible, and poisonous mushrooms, we assumed that none of the recruited students possessed certified competencies in recognizing macrofungi and assessing their edibility.

\subsection{Research tools}

To carry out the user study, we developed the following research tools: (1) an online survey questionnaire through which we collected, among other things: basic information about our users, especially about their experience in data analysis and visualization, as well as their domain knowledge about mushrooms, which was connected with the dataset we chose; (2) a presentation, with slides showing selected XAI techniques, to which we asked respondents questions and which respondents modified in terms of their subjective assessment of the usefulness of the individual slides and their components (similar to the well-known UX card sorting technique); (3) a set of questions and tasks used during the TAP; (4) a study scenario and other supporting materials, including an audio file with a recording of a sample verbalisation in the TAP and a test presentation to explain the procedure to respondents; (5) a simple UX questionnaire with which users could determine on a five-point scale their overall feelings related to the use of the viewed explanations in terms of their level of difficulty, aesthetic, comprehensibility and stimulation of the user's attention. Additionally, users were asked to indicate data visualizations that were easier and more difficult than those seen during the study, which was inspired by the Kano technique derived from market research aimed at showing the tested product in relation to selected aspects often by comparing it to other, similar ones; (6) consent forms and information for respondents.  

\subsubsection{Think-aloud protocol (TAP)}
Think-aloud protocol (TAP) is a data collection technique that has its roots in psychology and is based on the verbalisation of the subject's thoughts, feelings and impressions~\cite{someren1994tap}. This technique allows us to capture both the pragmatic values of UX (such as usability and functionality) as well as the subjective emotional and aesthetic impressions that accompany the user's interaction with an information product~\cite{zych2017ux}, e.g. a website, a database or, in our case, a presentation with XAI techniques. TAP has three variants: concurrent (verbalisation occurs at the same time as the activity under investigation), retrospective (verbalisation occurs only after the activity has taken place, for example in the form of commenting on a recording of an activity previously performed by the participant) and constructive interaction (another name: co-discovery learning) based on a conversation between two people~\cite{fan2019tap,VANDENHAAK2009tap}. In the study of XAI techniques, we used a concurrent variant of TAP, which does not overload the participants' memory mechanism and does not further prolong the study procedure.   

In general, the advantages of TAP include: its versatility (it is suitable for examining different elements of UX, can be used at different phases in the product development cycle, and can be applied in a quantitative, qualitative and mixed approach), mobility (TAP typically does not require specialised equipment, so it can be used in a variety of environmental conditions that are compatible with the scenario and purpose of the study in question), and low cost (which results from not needing specialised technical equipment or software). In addition, properly conducted TAP provides information on the actual impressions of users, thus providing convincing data.  With the right attitude on the part of the researcher (i.e. controlling his or her own behaviour so as not to impose his or her own opinions on the respondents and creating an atmosphere that encourages verbalisation), it is also a technique that is easy to use and learn.   

\subsubsection{Thematic analysis in MAXQDA}
In the conducted study, based on a qualitative methodology employing a think-aloud protocol, a thematic content analysis was also employed utilising inductive forms of reflexive analysis~\cite{braun2019tap,cisek2014}.
In the conducted study, based on a qualitative methodology employing a think-aloud protocol, a thematic content analysis was also employed utilising inductive forms of reflexive analysis~\cite{braun2019tap,cisek2014}.

The thematic analysis comprised stages such as: a) an exploratory phase, involving data reconnaissance and familiarization, b) inductive, independent generation of codes by two individuals, followed by c) possible deductive referencing to the main research goals of evaluation of explanations and recommendations, as well as potential theoretical references to quality information assessment criteria and their connections to XAI evaluation metrics (footnote), then d) creation and verification of themes, during which the first and subsequent codebooks were developed, and e) results compilation~\cite{braun2019tap}. 
A comprehensive list of codes used for thematic analysis was provided in the form of a codebook in the supplementary materials. 

\subsubsection{Analysis of the user experience}

The analysis of user experience was conducted by us with two independent approaches: 1) through analysis of the modifications that the users requested for the visualizations of XAI output, 2) through UX questionnaire and questions inspired by Kano approach. 

The analysis of modifications proposed by non-domain experts in the baseline presentation was based on a model of data presentation, combining data visualization and data storytelling~\cite{chao2019storytelling}. 
While both approaches serve to communicate data, they differ in their objectives. 
Data visualization primarily focuses on enhancing data perception, whereas data storytelling aims to transform data perception into data cognition by incorporating narrative elements~\cite{chao2019storytelling}.    

The study of modification of visualization was complemented by the UX questionnaire which is a popular technique for collecting information about a user's overall experience using the tested system, long described in the subject literature on UX research~\cite{Laugwitz2008ConstructionAE} and vigorously developed in relation to various situations and user groups~\cite{Schrepp2019DesignAV,wobbekind2021ux}.

\section{Data availability}

Additional data, such as summary of MAXQDA analysis and presentation slides that the participants were interacting with that we were referring throughout the text were provided in supplementary materials.
The source code used to reproduce the ML model and explanations is available in GitHub repository\footnote{\url{https://gitlab.geist.re/pro/xai-fungi/}}. 
Finally, we have prepared the whole dataset resulting from the experiment to be publicly available in the Zenodo platform~\cite{xaifungi}.

\section*{Contributions}
All authors contributed equally to the work;
G.J.N and M.W. contributed equally to the supervision of the work.

\section*{Acknowledgements}
The paper is funded from the XPM project funded by the National Science Centre, Poland under the CHIST-ERA programme grant agreement Np. 857925 (NCN UMO-2020/02/Y/ST6/00070).
The research has been supported by a grant from the Priority Research Area (DigiWorld) under the Strategic Programme Excellence Initiative at Jagiellonian University.

\bibliographystyle{ieeetr} 
\bibliography{manuscript-arxiv}

\begin{thebibliography}{10}

\bibitem{darpa}
DARPA, ``Explainable {Artificial} {Intelligence}.''
  https://www.darpa.mil/program/explainable-artificial-intelligence, 2015.

\bibitem{goodman2016regulations}
B.~Goodman and S.~Flaxman, ``{EU} regulations on algorithmic decision-making
  and a "right to explanation",'' 2016.
\newblock cite arxiv:1606.08813Comment: presented at 2016 ICML Workshop on
  Human Interpretability in Machine Learning (WHI 2016), New York, NY.

\bibitem{aiact2022hacker}
P.~Hacker and J.-H. Passoth, ``Varieties of {AI} {Explanations} {Under} the
  {Law}. {From} the {GDPR} to the {AIA}, and {Beyond},'' in {\em {xxAI} -
  {Beyond} {Explainable} {AI}: {International} {Workshop}, {Held} in
  {Conjunction} with {ICML} 2020, {July} 18, 2020, {Vienna}, {Austria},
  {Revised} and {Extended} {Papers}} (A.~Holzinger, R.~Goebel, R.~Fong,
  T.~Moon, K.-R. Müller, and W.~Samek, eds.), Lecture {Notes} in {Computer}
  {Science}, pp.~343--373, Cham: Springer International Publishing, 2022.

\bibitem{Longo_2024}
L.~Longo, M.~Brcic, F.~Cabitza, J.~Choi, R.~Confalonieri, J.~D. Ser,
  R.~Guidotti, Y.~Hayashi, F.~Herrera, A.~Holzinger, R.~Jiang, H.~Khosravi,
  F.~Lecue, G.~Malgieri, A.~Páez, W.~Samek, J.~Schneider, T.~Speith, and
  S.~Stumpf, ``Explainable artificial intelligence (xai) 2.0: A manifesto of
  open challenges and interdisciplinary research directions,'' {\em Information
  Fusion}, vol.~106, p.~102301, June 2024.

\bibitem{miller2019social}
T.~Miller, ``Explanation in artificial intelligence: Insights from the social
  sciences,'' {\em Artificial Intelligence}, vol.~267, pp.~1--38, 2019.

\bibitem{ghassemi2021xaimedpitfalls}
M.~Ghassemi, L.~Oakden-Rayner, and A.~L. Beam, ``The false hope of current
  approaches to explainable artificial intelligence in health care,'' {\em The
  Lancet Digital Health}, vol.~3, pp.~e745--e750, Nov. 2021.

\bibitem{verma2021industryxaipitfalls}
S.~Verma, A.~Lahiri, J.~P. Dickerson, and S.-I. Lee, ``Pitfalls of
  {Explainable} {ML}: {An} {Industry} {Perspective},'' June 2021.
\newblock arXiv:2106.07758 [cs].

\bibitem{roski2021xaimedfail}
J.~Roski, E.~J. Maier, K.~Vigilante, E.~A. Kane, and M.~E. Matheny,
  ``{Enhancing trust in AI through industry self-governance},'' {\em Journal of
  the American Medical Informatics Association}, vol.~28, pp.~1582--1590, 04
  2021.

\bibitem{evans2022xaifailpathology}
T.~Evans, C.~O. Retzlaff, C.~Geißler, M.~Kargl, M.~Plass, H.~Müller, T.-R.
  Kiehl, N.~Zerbe, and A.~Holzinger, ``The explainability paradox: {Challenges}
  for {xAI} in digital pathology,'' {\em Future Generation Computer Systems},
  vol.~133, pp.~281--296, Aug. 2022.

\bibitem{ehsan2021xaipitfalls}
U.~Ehsan and M.~O. Riedl, ``Explainability {Pitfalls}: {Beyond} {Dark}
  {Patterns} in {Explainable} {AI},'' Sept. 2021.
\newblock arXiv:2109.12480 [cs].

\bibitem{molnar2022pitfalls}
C.~Molnar, G.~König, J.~Herbinger, T.~Freiesleben, S.~Dandl, C.~A. Scholbeck,
  G.~Casalicchio, M.~Grosse-Wentrup, and B.~Bischl, ``General {Pitfalls}
  of {Model}-{Agnostic} {Interpretation} {Methods} for {Machine} {Learning}
  {Models},'' in {\em {xxAI} - {Beyond} {Explainable} {AI}: {International}
  {Workshop}, {Held} in {Conjunction} with {ICML} 2020, {July} 18, 2020,
  {Vienna}, {Austria}, {Revised} and {Extended} {Papers}} (A.~Holzinger,
  R.~Goebel, R.~Fong, T.~Moon, K.-R. Müller, and W.~Samek, eds.), Lecture
  {Notes} in {Computer} {Science}, pp.~39--68, Cham: Springer International
  Publishing, 2022.

\bibitem{xu2006relevance}
Y.~C. Xu and Z.~Chen, ``Relevance judgment: What do information users consider
  beyond topicality?,'' {\em Journal of the American Society for Information
  Science and Technology}, vol.~57, no.~7, pp.~961--973, 2006.

\bibitem{lee2012gratification}
C.~S. Lee and L.~Ma, ``News sharing in social media: The effect of
  gratifications and prior experience,'' {\em Computers in Human Behavior},
  vol.~28, no.~2, pp.~331--339, 2012.

\bibitem{SHAPlun}
S.~Lundberg and S.-I. Lee, ``A unified approach to interpreting model
  predictions,'' pp.~4765--4774, Curran Associates, Inc., 2017.

\bibitem{lime}
M.~T. Ribeiro, S.~Singh, and C.~Guestrin, ``“{W}hy should i trust you?”:
  Explaining the predictions of any classifier,'' in {\em Proceedings of the
  22nd ACM SIGKDD International Conference on Knowledge Discovery and Data
  Mining}, KDD ’16, (New York, NY, USA), p.~1135–1144, Association for
  Computing Machinery, 2016.

\bibitem{anchors}
M.~T. Ribeiro, S.~Singh, and C.~Guestrin, ``Anchors: high-precision
  model-agnostic explanations,'' in {\em Proceedings of the {Thirty}-{Second}
  {AAAI} {Conference} on {Artificial} {Intelligence} and {Thirtieth}
  {Innovative} {Applications} of {Artificial} {Intelligence} {Conference} and
  {Eighth} {AAAI} {Symposium} on {Educational} {Advances} in {Artificial}
  {Intelligence}}, {AAAI}'18/{IAAI}'18/{EAAI}'18, (New Orleans, Louisiana,
  USA), pp.~1527--1535, AAAI Press, Feb. 2018.

\bibitem{dice2020}
R.~K. Mothilal, A.~Sharma, and C.~Tan, ``Explaining machine learning
  classifiers through diverse counterfactual explanations,'' in {\em
  Proceedings of the 2020 Conference on Fairness, Accountability, and
  Transparency}, FAT* '20, (New York, NY, USA), p.~607–617, Association for
  Computing Machinery, 2020.

\bibitem{FromLocalExplanationsToGlobal}
S.~M. Lundberg, G.~Erion, H.~Chen, A.~DeGrave, J.~M. Prutkin, B.~Nair, R.~Katz,
  J.~Himmelfarb, N.~Bansal, and S.-I. Lee, ``From local explanations to global
  understanding with explainable ai for trees,'' {\em Nature Machine
  Intelligence}, vol.~2, pp.~56--67, Jan 2020.

\bibitem{baniecki2023grammar}
H.~Baniecki, D.~Parzych, and P.~Biecek, ``The grammar of interactive
  explanatory model analysis,'' {\em Data Mining and Knowledge Discovery},
  2023.

\bibitem{nielsentap}
N.~Norman, ``Thinking aloud: The \#1 usability tool,'' 2012.

\bibitem{puryear2016inside}
J.~S. Puryear, ``Inside the creative sifter: Recognizing metacognition in
  creativity development,'' vol.~50, no.~4, pp.~321--332, 2016.
\newblock Place: United Kingdom Publisher: Wiley-Blackwell Publishing Ltd.

\bibitem{josephsen2017qualitative}
J.~M. Josephsen, ``A qualitative analysis of metacognition in simulation,''
  vol.~56, no.~11, pp.~675--678, 2017.

\bibitem{rojas2020qualitative}
L.~A. Rojas~P., M.~E. Truyol, J.~F. Calderon~Maureira,
  M.~Orellana~Qui{\~{n}}ones, and A.~Puente, ``Qualitative evaluation of the
  usability of a web-based survey tool to assess reading comprehension and
  metacognitive strategies of university students,'' in {\em Social Computing
  and Social Media. Design, Ethics, User Behavior, and Social Network Analysis}
  (G.~Meiselwitz, ed.), (Cham), pp.~110--129, Springer International
  Publishing, 2020.

\bibitem{Mikuli2011ACR}
J.~Mikuli{\'c} and D.~Prebežac, ``A critical review of techniques for
  classifying quality attributes in the kano model,'' {\em Managing Service
  Quality}, vol.~21, pp.~46--66, 2011.

\bibitem{Laugwitz2008ConstructionAE}
B.~Laugwitz, T.~Held, and M.~Schrepp, ``Construction and evaluation of a user
  experience questionnaire,'' in {\em Symposium of the Workgroup Human-Computer
  Interaction and Usability Engineering of the Austrian Computer Society},
  2008.

\bibitem{atkin2023pierce}
A.~Atkin, ``{Peirce’s Theory of Signs},'' Metaphysics Research Lab, Stanford
  University, {S}pring 2023~ed., 2023.

\bibitem{someren1994tap}
M.~W. van Someren, Y.~F. Barnard, and J.~a.~C. Sandberg.
\newblock {LondenAcademic} Press, 1994.

\bibitem{zych2017ux}
M.~Zych, ``Przekaz symboliczny i podejście user experience na przykładzie
  serwisów internetowych teatrów krakowskich,'' vol.~55, no.~1, pp.~146--158,
  2017.
\newblock Number: 1(109).

\bibitem{fan2019tap}
M.~Fan, J.~Lin, C.~Chung, and K.~N. Truong, ``Concurrent think-aloud
  verbalizations and usability problems,'' {\em ACM Trans. Comput.-Hum.
  Interact.}, vol.~26, jul 2019.

\bibitem{VANDENHAAK2009tap}
M.~J. {van den Haak}, M.~D. {de Jong}, and P.~J. Schellens, ``Evaluating
  municipal websites: A methodological comparison of three think-aloud
  variants,'' {\em Government Information Quarterly}, vol.~26, no.~1,
  pp.~193--202, 2009.
\newblock From Implementation to Adoption: Challenges to Successful
  E-government Diffusion.

\bibitem{braun2019tap}
V.~Braun, V.~Clarke, N.~Hayfield, and G.~Terry, {\em Thematic Analysis},
  pp.~843--860.
\newblock Singapore: Springer Singapore, 2019.

\bibitem{cisek2014}
S.~Cisek, ``Analiza danych jakościowych we współczesnej informatologii,''
  (Warszawa), pp.~79--88, Wydawnictwo SBP, 2014.

\bibitem{chao2019storytelling}
L.~Chao and C.~Zhang, ``Data storytelling: From data perception to data
  cognition,'' {\em Journal of Library Science in China}, vol.~45, no.~5,
  pp.~61--78, 2019.

\bibitem{Schrepp2019DesignAV}
M.~Schrepp and J.~Thomaschewski, ``Design and validation of a framework for the
  creation of user experience questionnaires,'' {\em Int. J. Interact. Multim.
  Artif. Intell.}, vol.~5, pp.~88--95, 2019.

\bibitem{wobbekind2021ux}
L.~Wöbbekind, T.~Mandl, and C.~Womser-Hacker, ``Construction and first testing
  of the ux kids questionnaire (uxkq): A tool for measuring pupil's user
  experience in interactive learning apps using semantic differentials,'' in
  {\em Mensch und Computer 2021 - Tagungsband}, pp.~473--484, New York: ACM,
  2021.

\bibitem{xaifungi}
S.~Bobek, P.~Korycińska, M.~Krakowska, M.~Mozolewski, D.~Rak, M.~Zych,
  M.~Wójcik, and G.~J. Nalepa, ``{XAI-FUNGI: Dataset from the user study on
  comprehensibility of XAI algorithms},'' Oct. 2024.

\end{thebibliography}

\end{document}